\documentstyle[twocolumn]{article} 

\title{EVALUATING DISCOURSE PROCESSING ALGORITHMS \\
       (Appeared in ACL89, Vancouver)}
\author{Marilyn A. Walker}
\date{Hewlett Packard Laboratories  \\
      Filton Rd., Bristol, England BS12 6QZ, U.K.  \\
      \& University of Pennsylvania \\
      lyn\%lwalker@hplb.hpl.hp.com \\}

\input{psfig}
\begin{document}           
\bibliographystyle{alpha}  
\maketitle                 

\begin{abstract}

In order to take steps towards establishing a methodology for
evaluating Natural Language systems, we conducted a case study.  We
attempt to evaluate two different approaches to anaphoric processing
in discourse by comparing the accuracy and coverage of two published
algorithms for finding the co-specifiers of pronouns in naturally
occurring texts and dialogues.  We present the quantitative results of
hand-simulating these algorithms, but this analysis naturally gives
rise to both a qualitative evaluation and recommendations for
performing such evaluations in general.  We illustrate the general
difficulties encountered with quantitative evaluation.  These are
problems with: (a) allowing for underlying assumptions, (b)
determining how to handle underspecifications, and (c) evaluating the
contribution of false positives and error chaining.  \end{abstract}

\section{Introduction}

     In the course of developing natural language interfaces, computational
linguists are often
in the position of evaluating different theoretical approaches to the analysis
of natural
language (NL).  They might want to (a) evaluate and improve on a current
system, (b) add a
capability to a system that it didn't previously have, (c) combine modules from
different
systems.

Consider the goal of adding a discourse component to a system, or evaluating
and improving one
that is already in place.  A discourse module might combine theories on, e.g.,
centering or
local focusing \cite{GJW83,Sidner79}, global focus \cite{Grosz77}, coherence
relations\cite{Hobbs85a}, event reference \cite{Webber86}, intonational
structure \cite{PH90},
system vs. user beliefs \cite{Pollack86b}, plan or intent recognition or
production
\cite{Cohen78,AP80,SI81}, control\cite{WS88}, or complex syntactic structures
\cite{Prince85}. How might one evaluate the relative contributions of each of
these factors or
compare two approaches to the same problem?

In order to take steps towards establishing a methodology for doing this type
of comparison, we
conducted a case study.  We attempt to evaluate two different approaches to
anaphoric processing
in discourse by comparing the accuracy and coverage of two published algorithms
for finding the
co-specifiers of pronouns in naturally occurring texts and
dialogues\cite{Hobbs76a,BFP87}.  Thus
there are two parts to this paper: we present the quantitative results of
hand-simulating these
algorithms (henceforth Hobbs algorithm and BFP algorithm), but this analysis
naturally gives
rise to both a qualitative evaluation and recommendations for performing such
evaluations in
general.  We illustrate the general difficulties encountered with quantitative
evaluation.
These are problems with: (a) allowing for underlying assumptions, (b)
determining how to handle
underspecifications, and (c) evaluating the contribution of false positives and
error chaining.

Although both algorithms are part of theories of discourse that posit the
interaction of the
algorithm with an inference or intentional component, we will not use reasoning
in tandem with
the algorithm's operation. We have made this choice because we want to be able
to analyse the
performance of the algorithms across different domains. We focus on the
linguistic basis of
these approaches, using only selectional restrictions, so that our analysis is
independent of
the vagaries of a particular knowledge representation.  Thus what we are
evaluating is the
extent to which these algorithms suffice to narrow the search of an inference
component\footnote{But note the definition of success in section \ref{algs}.}.
This analysis
gives us some indication of the contribution of syntactic constraints, task
structure and global
focus to anaphoric processing.

The data on which we compare the algorithms are important if we are to evaluate
claims of
generality.  If we look at types of NL input, one clear division is between
textual and
interactive input. A related, though not identical factor is whether the
language being analysed
is produced by more than one person, although this distinction may be conflated
in textual
material such as novels that contain reported conversations. Within two-person
interactive
dialogues, there are the task-oriented master-slave type, where all the
expertise and hence much
of the initiative, rests with one person. In other two-person dialogues, both
parties may
contribute discourse entities to the conversation on a more equal basis.  Other
factors of
interest are whether the dialogues are human-to-human or human-to-computer, as
well as the
modality of communication, e.g. spoken or typed, since some researchers have
indicated that
dialogues, and particularly uses of reference within them, vary along these
dimensions
\cite{Cohen84a,Thom80,Gui86,DJ89,WS89}.

We analyse the performance of the algorithms on three types of data.  Two of
the samples are
those that Hobbs used when developing his algorithm. One is an excerpt from a
novel and the
other a sample of journalistic writing.  The remaining sample is a set of 5
human-human,
keyboard-mediated, task-oriented dialogues about the assembly of a plastic
water pump
\cite{Cohen84a}. This covers only a subset of the above types.  Obviously it
would be instructive
to conduct a similar analysis on other textual types.

\section{Quantitative Evaluation--Black Box}
\subsection{The Algorithms}
\label{algs}

When embarking on such a comparison, it would be convenient to assume that the
inputs to the
algorithms are identical and compare their outputs.  Unfortunately since
researchers do not even
agree on which phenomena can be explained syntactically and which semantically,
the boundaries
between two modules are rarely the same in NL systems.  In this case the BFP
centering algorithm
and Hobbs algorithm both make {\sc assumptions} about other system components.
These are, in
some sense, a further specification of the operation of the algorithms that
must be made in
order to hand-simulate the algorithms.  There are two major sets of
assumptions, based on
discourse segmentation and syntactic representation. We attempt to make these
explicit for each
algorithm and pinpoint where the algorithms might behave differently were these
assumptions not
well-founded.

In addition, there may be a number of {\sc underspecifications} in the
descriptions of the
algorithms. These often arise because theories that attempt to categorize
naturally occurring
data and algorithms based on them will always be prey to previously
unencountered examples.  For
example, since the BFP salience hierarchy for discourse entities is based on
grammatical
relation, an implicit assumption is that an utterance only has one subject.
However the novel
{\sl Wheels} has many examples of reported dialogue such as {\em She continued,
unperturbed,
``Mr. Vale quotes the Bible about air pollution.'' } One might wonder whether
the subject is
{\em She} or {\em Mr. Vale}. In some cases, the algorithm might need to be
further specificied
in order to be able to process any of the data, whereas in others they may just
highlight where
the algorithm needs to be modified (see section \ref{mod}). In general we count
underspecifications as failures.

Finally, it may not be clear what the {\sc definition of success} is.  In
particular it is not
clear what to do in those cases where an algorithm produces multiple or partial
interpretations.
In this situation a system might flag the utterance as ambiguous and draw in
support from other
discourse components. This arises in the present analysis for two reasons: (1)
the constraints
given by \cite{GJW86} do not always allow one to choose a preferred
interpretation, (2) the BFP
algorithm proposes equally ranked interpretations in parallel.  This doesn't
happen with the
Hobbs algorithm because it proposes interpretations in a sequential manner, one
at a time.  We
chose to count as a failure those situations in which the BFP algorithm only
reduces the number
of possible interpretations, but Hobbs algorithm stops with a correct
interpretation.  This
ignores the fact that Hobbs may have rejected a number of interpretations
before stopping.  We
also have not needed to make a decision on how to score an algorithm that only
finds one
interpretation for an utterance that humans find ambiguous.

\subsubsection{Centering algorithm}
The centering algorithm as defined by Brennan, Friedman and Pollard, (BFP
algorithm), is derived
from a set of rules and constraints put forth by Grosz, Joshi and Weinstein
\cite{GJW83,GJW86}.
We shall not reproduce this algorithm here (See \cite{BFP87}).  There are two
main structures in
the centering algorithm, the {\sc Cb}, the {\sc backward looking center}, which
is what the
discourse is `about', and an ordered list, {\sc Cf}, of {\sc forward looking
centers}, which are
the discourse entities available to the next utterance for pronominalization.
The centering
framework predicts that in a local coherent stretch of dialogue, speakers will
prefer to {\sc
continue} talking about the same discourse entity, that the {\sc Cb} will be
the highest ranked
entity of the previous utterance's forward centers that is realized in the
current utterance,
and that if anything is pronominalized the {\sc Cb} must be.

In the centering framework, the order of the forward-centers list is intended
to reflect the
salience of discourse entities. The BFP algorithm orders this list by
grammatical relation of
the complements of the main verb, i.e.  first the subject, then object, then
indirect object,
then other subcategorized-for complements, then noun phrases found in adjunct
clauses.  This
captures the intuition that subjects are more salient than other discourse
entities.

The BFP algorithm added linguistic constraints on {\sc contra-indexing} to the
centering
framework.  These constraints are exemplified by the fact that, in the sentence
{\em he likes
him}, the entity cospecified by {\em he} cannot be the same as that cospecified
by {\em him}. We
say that {\em he} and {\em him} are {\sc contra-indexed}.  The BFP algorithm
depends on semantic
processing to precompute these constraints, since they are derived from the
syntactic structure,
and depend on some notion of c-command\cite{Reinhart76}.  The other assumption
that is dependent
on syntax is that the the representations of discourse entities can be marked
with the
grammatical function through which they were realized, e.g.  subject.

     The BFP algorithm assumes that some other mechanism can structure both
written texts and
task-oriented dialogues into hierarchical segments. The present concern is not
with whether
there might be a grammar of discourse that determines this structure, or
whether it is derived
from the cues that cooperative speakers give hearers to aid in processing.
Since centering is a
local phenomenon and is intended to operate within a segment, we needed to
deduce a segmental
structure in order to analyse the data.  Speaker's intentions, task structure,
cue words like
{\em O.K.  now..}, intonational properties of utterances, coherence relations,
the scoping of
modal operators, and mechanisms for shifting control between discourse
participants have all
been proposed as ways of determining discourse segmentation
\cite{Grosz77,GS86,Reichman85,PH90,HL87,Hobbs78,Hobbs85a,Roberts88,WS88}.
Here, we use a combination
of orthography, anaphora distribution, cue words and task structure. The rules
are:
\begin{itemize}
\item
In published texts, a paragraph is a new segment unless the first sentence has
a pronoun in subject position or a pronoun where none of the preceding
sentence-internal noun phrases match its syntactic features.
\item
In the task-oriented dialogues, the action {\sc Pick-up} marks task
boundaries hence segment boundaries. Cue words like {\em next, then,} and {\em
now} also mark segment boundaries.  These will usually co-occur but either one
is sufficient for marking a segment boundary.
\end{itemize}

BFP never state that cospecifiers for pronouns within the same segment are
preferred over those
in previous segments, but this is an implicit assumption, since this line of
research is derived
from Sidner's work on local focusing. Segment initial utterances therefore are
the only
situation where the BFP algorithm will prefer a within-sentence noun phrase as
the cospecifier
of a pronoun.

\subsubsection{Hobbs' algorithm}

The Hobbs algorithm is based on searching for a pronoun's co-specifier in the
syntactic parse
tree of input sentences \cite{Hobbs76a}. We reproduce this algorithm in full in
the appendix
along with an example.  Hobbs algorithm operates on one sentence at a time, but
the structure of
previous sentences in the discourse is available.  It is stated in terms of
searches on parse
trees. When looking for an intrasentential antecedent, these searches are
conducted in a
left-to-right, breadth-first manner.  However, when looking for a pronoun's
antecedent within a
sentence, it will go sequentially further and further up the tree to the left
of the pronoun,
and that failing will look in the previous sentence. Hobbs does not assume a
segmentation of
discourse structure in this algorithm; the algorithm will go back arbitrarily
far in the text to
find an antecedent. In more recent work, Hobbs uses the notion of {\sc
coherence relations}
to structure the discourse \cite{HM87}.

The order by which Hobbs' algorithm traverses the parse tree is the closest
thing in his
framework to predictions about which discourse entities are salient. In the
main it prefers
co-specifiers for pronouns that are within the same sentence, and also ones
that are closer to
the pronoun in the sentence.  This amounts to a claim that different discourse
entities are
salient, depending on the position of a pronoun in a sentence. When seeking an
intersentential
co-specification, Hobbs algorithm searches the parse tree of the previous
utterance
breadth-first, from left to right.  This predicts that entities realized in
subject position are
more salient, since even if an adjunct clause linearly precedes the main
subject, any noun
phrases within it will be deeper in the parse tree.  This also means that
objects and indirect
objects will be among the first possible antecedents found, and in general that
the depth of
syntactic embedding is an important determiner of discourse prominence.

Turning to the assumptions about syntax, we note that Hobbs assumes that one
can produce the
correct syntactic structure for an utterance, with all adjunct phrases attached
at the proper
point of the parse tree.  In addition, in order to obey linguistic constraints
on coreference,
the algorithm depends on the existence of a $\overline{\rm{N}}$ parse tree
node, which denotes a
noun phrase without its determiner (See the example in the Appendix).  Hobbs
algorithm
procedurally encodes contra-indexing constraints by skipping over NP nodes
whose
$\overline{\rm{N}}$ node dominates the part of the parse tree in which the
pronoun is found,
which means that he cannot guarantee that two contra-indexed pronouns will not
choose the same
NP as a co-specifier.

Hobbs also assumes that his algorithm can somehow collect discourse entities
mentioned
alone into sets as co-specifiers of plural anaphors.
Hobbs discusses at length other assumptions that he makes about the
capabilities of an
interpretive process that operates before the algorithm \cite{Hobbs76a}.  This
includes such
things as being able to recover syntactically recoverable omitted text, such as
elided verb
phrases, and the identities of the speakers and hearers in a dialogue.

\subsubsection{Summary}
\label{sum}

A major component of any discourse algorithm is the prediction of which
entities are salient,
even though all the factors that contribute to the salience of a discourse
entity have not been
identified \cite{Prince81,Prince85,BF83,HTD86}. So an obvious question is when
the two
algorithms actually make different predictions. The main difference is that the
choice of a
co-specifier for a pronoun in the Hobbs algorithm depends in part on the
position of that
pronoun in the sentence. In the centering framework, no matter what criteria
one uses to order
the forward-centers list, pronouns take the most salient entities as
antecedents, irrespective
of that pronoun's position.  Hobbs ordering of entities from a previous
utterance varies from
BFP in that possessors come before case-marked objects and indirect objects,
and there may
be some other differences as well but none of them were relevant to the
analysis that follows.

The effects of some of the assumptions are measurable and we will attempt to
specify exactly
what these effects are, however some are not, e.g.  we cannot measure the
effect of Hobbs'
syntax assumption since it is difficult to say how likely one is to get the
wrong parse.
We adopt the set collection assumption for both algorithms as well as the
ability
to recover the identity of speakers and hearers in dialogue.

\subsection{Quantitative Results of the Algorithms}
\label{perf-sec}

The texts on which the algorithms are analysed are the first chapter of Arthur
Hailey's novel
{\sl Wheels}, and the July 7, 1975 edition of Newsweek.  The sentences in {\sl
Wheels} are short
and simple with long sequences consisting of reported conversation, so it is
similar to a
conversational text. The articles from {\sl Newsweek} are typical of
journalistic writing.  For
each text, the first 100 occurrences of singular and plural third-person
pronouns were used to
test the performance of the algorithms.  The task-dialogues contain a total of
81 uses of {\em
it} and no other pronouns except for {\em I} and {\em you}.  In the figures
below note that
possessives like {\em his} are counted along with {\em he} and that accusatives
like {\em him}
and {\em her} are counted as {\em he} and {\em she}\footnote{Hobbs reports his
algorithm's
performance and the examples it fails on in \cite{Hobbs76a,Hobbs76b}. The
numbers reported here
vary slightly from those. This is probably due to a discrepancy in exactly what
the data-set
consisted of.}.

\begin{figure}[htb]
\centering
\begin{tabular}{p{.75in}c||cc}
& N & Hobbs & BFP \\ \hline
Wheels & 100 & 88 & 90 \\
Newsweek & 100 & 89 & 79 \\
Tasks  & 81 & 51 & 49  \\ \hline
\end{tabular}
\caption{Number correct for both algorithms for
Wheels, Newsweek and Task Dialogues}
\label{res-fig}
\end{figure}

We performed three analyses on the quantitative results.  A comparison of the
two algorithms on
each data set individually and an overall analysis on the three data sets
combined revealed no
significant differences in the performance of the two algorithms ($\chi^2 =
3.25$, not
significant). In addition for each algorithm alone we tested whether there were
significant
differences in performance for different textual types. Both of the algorithms
performed
significantly worse on the task dialogues ($\chi^2 = 22.05$ for Hobbs, $\chi^2
= 21.55$ for BFP,
$p < 0.05$).

We might wonder with what confidence we should view these numbers. A
significant factor that
must be considered is the contribution of {\sc false positives} and {\sc error
chaining}.  A
{\sc false positive} is when an algorithm gets the right answer for the wrong
reason. A very
simple example of this phenomena is illustrated by this sequence from one of
the task dialogues.

\begin{tabular}{ll}
$\rm{Exp_1}$: & Now put IT in the pan of water. \\
$\rm{Exp_2}$: & Stand IT up. \\
$\rm{Exp_3}$: & Pump the little handle with the red cap \\
    &  on IT. \\
$\rm{Cli_1}$. & ok \\
$\rm{Exp_4}$. & Does IT work?? \\
\end{tabular}

The first {\em it} in $\rm{Exp_1}$ refers to {\em the pump}.  Hobbs algorithm
gets the right
antecedent for {\em it} in $\rm{Exp_3}$, which is {\em the little handle}, but
then fails on {\em it}
in $\rm{Exp_4}$, whereas the BFP algorithm has {\em the pump} centered at
$\rm{Exp_1}$ and continues to
select that as the antecedent for {\em it} throughout the text. This means BFP
gets the wrong
co-specifier in $\rm{Exp_3}$ but this error allows it to get the correct
co-specifier in $\rm{Exp_4}$.

Another type of false positive example is {\em ``Everybody and HIS brother
suddenly wants to be
the President's friend,'' said one aide.} Hobbs gets this correct as long as
one is willing to
accept that {\em Everybody} is really the antecedent of {\em his}. It seems to
me that this
might be an idiomatic use.

{\sc Error chaining} refers to the fact that once an algorithm makes an error,
other errors can
result.  Consider:

\begin{tabular}{ll}
$\rm{Cli_1}$: &Sorry no luck. \\
$\rm{Exp_1}$: &I bet IT's the stupid red thing.\\
$\rm{Exp_2}$: &Take IT out. \\
$\rm{Cli_2}$: &Ok. IT is stuck. \\
\end{tabular}

In this example once an algorithm fails at $\rm{Exp_1}$ it will fail on
$\rm{Exp_2}$ and
$\rm{Cli_2}$ as well since the choices of a cospecifier in the following
examples
are dependent on the choice in $\rm{Exp_1}$.

It isn't possible to measure the effect of false positives, since in some sense
they are
subjective judgements. However one can and should measure the effects of error
chaining,
since reporting numbers that correct for error chaining is misleading, but if
the
error that produced the error chain can be corrected then the algorithm might
show
a significant improvement. In this analysis, error chains contributed 22
failures
to Hobbs' algorithm and 19 failures to BFP.

\section{Qualitative Evaluation--Glass Box}

The numbers presented in the previous section are intuitively unsatisfying.
They tell us nothing
about what makes the algorithms more or less general, or how they might be
improved.  In
addition, given the assumptions that we needed to make in order to produce
them, one might
wonder to what extent the data is a result of these assumptions. Figure
\ref{res-fig} also fails
to indicate whether the two algorithms missed the same examples or are covering
a different set
of phenomena, i.e.  what the relative distribution of the successes and
failures are.  But
having done the hand-simulation in order to produce such numbers, all of this
information is
available. In this section we will first discuss the relative importance of
various factors that
go into producing the numbers above, then discuss if the algorithms can be
modified since the
flexibility of a framework in allowing one to make modifications is an
important dimension of
evaluation.

\subsection{Distributions}

The figures \ref{wheels-fig2}, \ref{news-fig2} and \ref{task-fig2} show for
each pronominal
category, the distribution of successes and failures for both algorithms.

\begin{figure}[htb]
\centering
\begin{tabular}{p{.5in}||cccc}
& Both & Neither & Hobbs & BFP \\
& & & only & only \\ \hline
HE & 66 & 1 & 1 & 7 \\
SHE  & 6 &  & & \\
IT & 6 & 3 & 3 & \\
THEY & 5 & 1 & 1 &  \\ \hline
Total & 83 & 5 & 5 & 7 \\
\end{tabular}
\caption{Distribution on Wheels}
\label{wheels-fig2}
\end{figure}

\begin{figure}[htb]
\centering
\begin{tabular}{p{.5in}||cccc}
& Both & Neither & Hobbs & BFP \\
& & & only & only \\ \hline
HE & 53 & & 8 & 2\\
IT & 11 & 5 & 4 & 1 \\
THEY & 13 & 3 &  &   \\ \hline
Total & 77 & 8 & 12 & 3 \\
\end{tabular}
\caption{Distribution on Newsweek}
\label{news-fig2}
\end{figure}

\begin{figure}[htb]
\centering
\begin{tabular}{p{.5in}||cccc}
& Both & Neither & Hobbs & BFP \\
& & & only & only \\ \hline
IT & 48 & 29 & 3 & 1  \\ \hline
\end{tabular}
\caption{Distribution on Task Dialogues}
\label{task-fig2}
\end{figure}

Since the main purpose of evaluation must be to improve the theory that we are
evaluating, the
most interesting cases are the ones on which the algorithms' performance varies
and those that
neither algorithm gets correct. We discuss these below.

\subsubsection{Both}

In the {\sl Wheels} data, 4 examples rest on the assumption that the identities
of speakers and
hearers is recoverable. For example in {\em The GM president smiled.  ``Except
Henry will be
damned forceful and the papers won't print all HIS language.''}, getting the
{\em his} correct
here depends on knowing that it is the GM president speaking.  Only 4 examples
rest on being
able to produce collections or discourse entities, and 2 of these occurred with
an explicit
instruction to the hearer to produce such a collection by using the phrase {\em
them both}.

\subsubsection{Hobbs only}

There are 21 cases that Hobbs gets that BFP don't, and of these these a few
classes stand out.
In every case the relevant factor is Hobbs' preference for intrasentential
co-specifiers.

One class, $(n=3)$, is exemplified by {\em Put the little black ring into the
the large blue
CAP with the hole in IT.} All three involved using the preposition {\em with}
in a descriptive
adjunct on a noun phrase.  It may be that with-adjuncts are common in visual
descriptions, since
they were only found in our data in the task dialogues, and a quick inspection
of Grosz's
task-oriented dialogues revealed some as well\cite{Deutsch74}.

Another class, $(n=7)$, are possessives. In some cases the possessive
co-specified with the
subject of the sentence, e.g. {\em The SENATE took time from ITS paralyzing New
Hampshire
election debate to vote agreement}, and in others it was within a relative
clause and
co-specified with the subject of that clause, e.g. {\em The auto industry
should be able to
produce a totally safe, defect-free CAR that doesn't pollute ITS environment}.

Other cases seem to be syntactically marked subject matching with constructions
that link two S
clauses $(n=8)$.  These are uses of {\em more-than} in e.g. {\em but
Chamberlain grossed about
\$8.3 million more than HE could have made by selling on the home front}. There
also are S-if-S
cases, as in {\em Mondale said: ``I think THE MAFIA would be broke if IT
conducted all its
business that way.''} We also have subject matching in AS-AS examples as in
{\em ... and the
resulting EXPOSURE to daylight has become as uncomfortable as IT was
unaccustomed}, as well as
in sentential complements, such as {\em But another liberal, Minnesota's Walter
MONDALE, said HE %
had found a lot of incompetence in the agency's operations}.  The fact that
quite a few of these
are also marked with {\em But} may be significant.

In terms of the possible effects that we noted earlier, the {\sc definition of
success} (see
section \ref{algs}) favors Hobbs $(n=2)$.  Consider: \nopagebreak
\begin{quote}
K: Next take the red piece that is the smallest and insert it into the
hole in the side of the large plastic tube.
IT goes
in the hole nearest the end with the engravings on IT.
\end{quote}

The Hobbs algorithm will correctly choose {\em the end} as the antecedent for
the second {\em
it}. The BFP algorithm on the other hand will get two interpretations, one in
which the second
{\em it} co-specifies {\em the red piece} and one in which it co-specifies {\em
the end}. They
are both {\sc continuing} interpretations since the first {\em it} co-specifies
the {\sc Cb},
but the constraints don't make a choice.

\subsubsection{BFP only}

All of the examples on which BFP succeed and Hobbs fails have to do
with extended discussion of one discourse entity. For instance:

\begin{tabular}{ll}
$\rm{Exp_1}$: & Now take the blue cap with the two  \\
    & prongs sticking out ({\sc Cb} = blue cap) \\
$\rm{Exp_2}$: & and fit the little piece of pink plastic on IT.\\
    & Ok?  ({\sc Cb}= blue cap) \\
$\rm{Cli_1}$: & ok.  \\
$\rm{Exp_3}$: & Insert the rubber ring into that blue cap.  \\
    & ({\sc Cb}= blue cap)  \\
$\rm{Exp_4}$: & Now screw IT onto the cylinder. \\
\end{tabular}

On this example, Hobbs fails by choosing the co-specifier of {\em it} in
$\rm{Exp_4}$ to be {\em the
rubber ring}, even though the whole segment has been about {\em the blue cap}.

Another example from the novel {\sl WHEELS} is given below.  On this one Hobbs
gets the first use of {\em he} but then misses the next four, as a result of
missing the second one by choosing {\em a housekeeper} as the co-specifier for
{\em HIS}.
\begin{quote}
..An executive vice-president of Ford was preparing to leave for Detroit
Metropolitan Airport.
HE had already breakfasted, alone. A housekeeper had brought a tray to HIS desk
in the softly
lighted study where, since 5 a.m., HE had been alternately reading memoranda
(mostly on special
blue stationery which Ford vice-presidents used in implementing policy) and
dictating crisp
instructions into a recording machine. HE had scarcely looked up, either as the
mail arrived,
or while eating, as HE accomplished in an hour what would have taken...
\end{quote}

Since {\em an executive vice-president} is centered in the first sentence, and
continued in each
following sentence, the BFP algorithm will correctly choose the cospecifier.

\subsubsection{Neither}
\label{neither}

Among the examples that neither algorithm gets correctly are 20 examples from
the task dialogues
of {\em it} referring to the global focus, the pump. In 15 cases, these shifts
to global focus
are marked syntactically with a cue word such as {\em Now}, and are not marked
in 5 cases.
Presumably they are felicitous since the pump is visually salient.  Besides the
global focus
cases, pronominal references to entities that were not linguistically
introduced are rare.  The
only other example is an implicit reference to `the problem' of the pump not
working:

\begin{tabular}{ll}
$\rm{Cli_1}$: & Sorry no luck. \\
$\rm{Exp_1}$: & I bet IT's the stupid red thing.\\
\end{tabular}

We have only two examples of sentential or VP anaphora altogether, such as {\em
Madam
Chairwoman, said Colby at last, I am trying to run a secret intelligence
service. IT was a
forlorn hope.\/} Neither Hobbs algorithm nor BFP attempt to cover these
examples.

Three of the examples are uses of {\em it} that seem to be lexicalized with
certain verbs, e.g.
{\em They hit IT off real well}.  One can imagine these being treated as
phrasal lexical items,
and therefore not handled by an anaphoric processing component\cite{AS89}.

Most of the interchanges in the task dialogues consist of the client responding
to commands with
cues such as {\em O.K.} or {\em Ready} to let the expert know when they have
completed a task.
When both parties contribute discourse entities to the common ground, both
algorithms may fail
$(n=4)$.

Consider: \nopagebreak

\begin{tabular}{ll}
$\rm{Exp_1}$: & Now we have a little red piece left \\
$\rm{Exp_2}$: & and I don't know what to do with IT.  \\
$\rm{Cli_1}$: & Well, there is a hole in the green plunger \\
    & inside the cylinder. \\
$\rm{Exp_3}$: & I don't think IT goes in THERE. \\
$\rm{Exp_4}$: & I think IT may belong in the blue cap \\
    & onto which you put the pink piece \\
    & of plastic. \\
\end{tabular}

In $\rm{Exp_3}$, one might claim that {\em it} and {\em there} are
contraindexed, and that {\em
there} can be properly resolved to {\em a hole}, so that {\em it} cannot be any
of the noun
phrases in the prepositional phrases that modify {\em a hole}, but whether any
theory of
contra-indexing actually give us this is questionable.

The main factor seems to be that even though $\rm{Exp_1}$ is not syntactically
a question, {\em
the little red piece} is the focus of a question, and as such is in focus
despite the fact that
the syntactic construction {\em there is} supposedly focuses {\em a hole in the
green plunger
...}\cite{Sidner79}. These examples suggest that a questioned entity is left
focused until the
point in the dialogue at which the question is resolved. The fact that {\em
well} has been noted
as a marker of response to questions supports this analysis\cite{Schiffrin87}.
Thus the relevant
factor here may be the switching of control among discourse participants
\cite{WS88}.  These
mixed-initiative features make these sequences inherently different than text.

\subsection{Modifiability}
\label{mod}

Task structure in the pump dialogues is an important factor especially as it
relates to the use
of global focus. Twenty of the cases on which both algorithms fail are
references to {\em the
pump}, which is the global focus. We can include a global focus in the
centering framework, as a
separate notion from the current {\sc Cb}. This means that in the 15 out of 20
cases where the
shift to global focus is identifiably marked with a cue-word such as {\em now},
the segment
rules will allow BFP to get the global focus examples.

BFP can add the VP and the S onto the end of the forward centers list, as
Sidner does in her
algorithm for local focusing \cite{Sidner79}. This lets BFP get the two
examples of event
anaphora. Hobbs discusses the fact that his algorithm cannot be modified to get
event anaphora
in \cite{Hobbs76a}.

Another interesting fact is that in every case in which Hobbs' algorithm gets
the correct co-specifier and BFP didn't, the relevant factor is Hobbs'
preference for intrasentential co-specifiers. One view on these cases may be
that these are not discourse anaphora, but there seems to be no principled way
to make this distinction. However, Carter has proposed some extensions to
Sidner's algorithm for local focusing that seem to be relevant here(chap. 6,
\cite{Carter87}). He argues that intra-sentential candidates (ISCs) should
be preferred over candidates from the previous utterance, ONLY in the cases
where no discourse center has been established or the discourse center is
rejected for syntactic or selectional reasons. He then uses Hobbs algorithm
to produce an ordering of these ISCs. This is compatible with the centering
framework since it is underspecified as to whether one should always choose
to establish a discourse center with a co-specifier from a previous
utterance. If we adopt Carter's rule into the centering framework, we find
that of the 21 cases that Hobbs gets that BFP don't, in 7 cases there is no
discourse center established, and in another 4 the current center can be
rejected on the basis of syntactic or sortal information. Of these Carter's
rule clearly gets 5, and another 3 seem to rest on whether one might want
to establish a discourse entity from a previous utterance.  Since the
addition of this constraint does not allow BFP to get any examples that
neither algorithm got, it seems that this combination is a way of making
the best out of both algorithms.

The addition of these modifications changes the quantitative results. See the
Figure
\ref{res-fig2}.
\begin{figure}[htb]
\centering
\begin{tabular}{p{.75in}c||cc}
& N & Hobbs & BFP \\ \hline
Wheels & 100 & 88 & 93 \\
Newsweek & 100 & 89 & 84 \\
Tasks  & 81 & 51 & 64  \\ \hline
\end{tabular}
\caption{Number correct for both algorithms after Modifications, for
Wheels, Newsweek and Task Dialogues }
\label{res-fig2}
\end{figure}

However, the statistical analyses still show that there is no significant
difference in the performance of the algorithms in general. It is also still
the case that the
performance of each algorithm significantly varies depending on the data. The
only significant
difference as a result of the modifications is that the BFP algorithm now
performs significantly
better on the pump dialogues alone ($\chi^2 = 4.31, p < .05$).

\section{Conclusion}

We can benefit in two ways from performing such evaluations: (a) we get general
results on a
methodology for doing evaluation, (b) we discover ways we can improve current
theories.  A split
of evaluation efforts into quantitative versus qualitative is incoherent. We
cannot trust the
results of a quantitative evaluation without doing a considerable amount of
qualitative analyses
and we should perform our qualitative analyses on those components that make a
significant
contribution to the quantitative results; we need to be able to measure the
effect of various
factors. These measurements must be made by doing comparisons at the data
level.

In terms of general results, we have identified some factors that make
evaluations of this type
more complicated and which might lead us to evaluate solely quantitative
results with care.
These are: (a) To decide how to evaluate {\sc underspecifications} and the
contribution of {\sc
assumptions}, and (b) To determine the effects of {\sc false positives} and
{\sc error chaining}.
We advocate an approach in which the contribution of each underspecification
and assumption is
tabulated as well as the effect of error chains. If a principled way could be
found to
identify false positives, their effect should be reported as well as part of
any quantitative
evaluation.

In addition, we have taken a few steps towards determining the relative
importance of different
factors to the successful operation of discourse modules.  The percent of
successes that {\bf
both} algorithms get indicates that syntax has a strong influence, and that at
the very least we
can reduce the amount of inference required.  In 59\% to 82\% of the cases both
algorithms get
the correct result. This probably means that in a large number of cases there
was no potential
conflict of co-specifiers. In addition, this analysis has shown, that at least
for task-oriented
dialogues global focus is a significant factor, and in general discourse
structure is more
important in the task dialogues. However simple devices such as cue words may
go a long way
toward determining this structure.

Finally, we should note that doing evaluations such as this allows us to
determine the {\sc
generality} of our approaches. Since the performance of both Hobbs and BFP
varies according to the
type of the text, and in fact was significantly worse on the task dialogues
than on the texts,
we might question how their performance would vary on other inputs.  An
annotated corpus
comprising some of the various NL input types such as those I discussed in the
introduction
would go a long way towards giving us a basis against which we could evaluate
the generality of
our theories.

\section{Acknowledgements}
David Carter, Phil Cohen, Nick Haddock, Jerry Hobbs, Aravind Joshi, Don Knuth,
Candy Sidner,
Phil Stenton, Bonnie Webber, and Steve Whittaker have provided valuable
insights toward this
endeavor and critical comments on a multiplicity of earlier versions of this
paper.  Steve
Whittaker advised me on the statistical analyses.  I would like to thank Jerry
Hobbs for
encouraging me to do this in the first place.


\newpage
\appendix
\section{The Hobbs algorithm}
\label{hobbs}

     The algorithm and an example is reproduced below. In it, NP denotes {\sc
noun phrase} and S
denotes {\sc sentence}.

\begin{enumerate}
 \item
   Begin at the NP node immediately dominating the pronoun in the parse tree of
S.
 \item
      Go up the tree until you encounter an NP or S node. Call this node $X$,
and call the path used to reach it $p$.
 \item
     Traverse all branches below node $X$ to the left of path $p$ in a
left-to-right breadth-first fashion. Propose as the antecedent any NP node
encountered that has an NP or S node on the path from it to $X$.
\item
     If $X$ is not the highest S node in the sentence, continue to
step 5. Otherwise traverse the surface parse trees of previous
sentences in the text in reverse chronological order until an
acceptable antecedent is found; each tree is traversed in a
left-to-right, breadth-first manner, and when an NP node is
encountered, it is proposed as the antecedent.
\item
     From node $X$, go up the tree to the first NP or S node encountered. Call
this new node $X$, and call the path traversed to reach it $p$.
\item
     If $X$ is an NP node and if the path $p$ to $X$ did not pass through the
$\overline{\rm{N}}$
node that $X$ immediately dominates, propose $X$ as the antecedent.
\item
     Traverse all branches below node $X$ to the left of path $p$ in a
left-to-right, breadth-first manner, but do not go below any NP or S node
encountered.  Propose any NP or S node encountered as the antecedent.
\item
Go to step 4.
\end{enumerate}

     The purpose of steps 2 and 3 is to observe the contra-indexing
constraints.  Let us consider a simple conversational sequence.

\begin{tabular} {ll}
$\rm{U_1}$: &Lyn's mom is a gardener.\\
$\rm{U_2}$: &Craige likes her.
\end{tabular}

     We are trying to find the antecedent for {\em her} in the second
utterance.  Let us go through the algorithm step by step, using the
parse trees for $\rm{U_1}$ and $\rm{U_2}$ in the figure.

\begin{figure}[tb]
\centerline{\psfig{figure=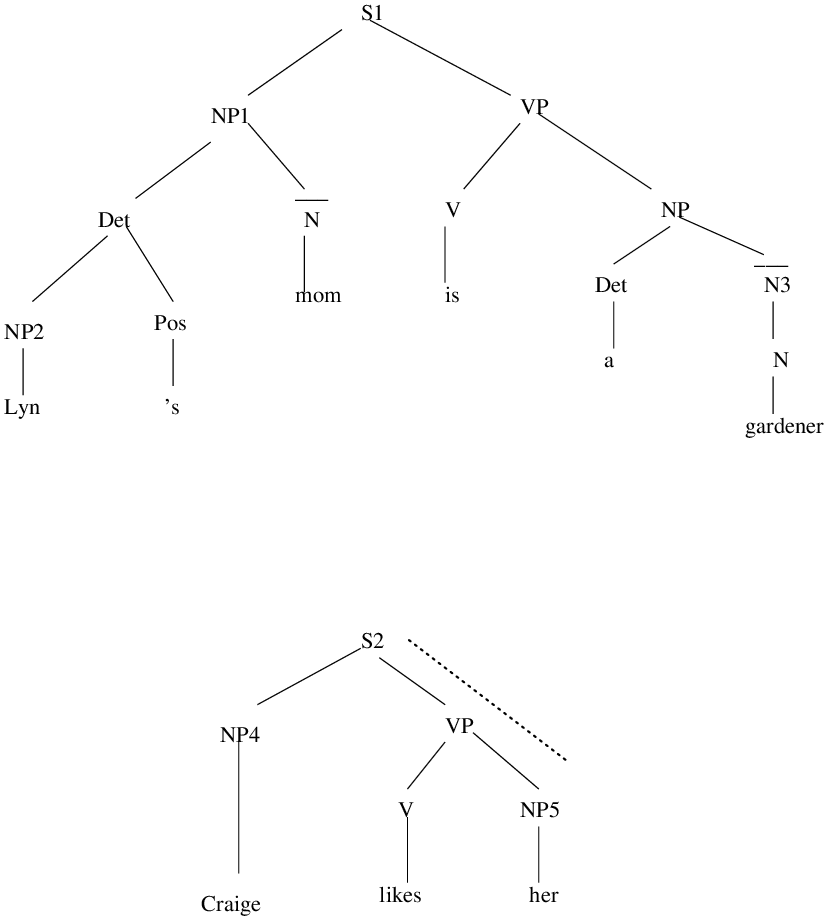,height=4.0in,width=3.0in}}
\label{Hobbs-fig}
\caption{Parse Trees for $\rm{U_1}$ and $\rm{U_2}$}
\end{figure}

\begin{enumerate}

 \item
    $\rm{NP_5}$ labels the starting point of step 1.

 \item
    $\rm{S_2}$ is called $X$. We mark the path $p$ with a dotted line.

 \item
     We traverse $\rm{S_2}$ to the left of $p$. We encounter $\rm{NP_4}$ but it
does not have
     an NP or S node between it and $X$. This means that $\rm{NP_4}$ is
contra-indexed with $\rm{NP_5}$.
     Note that if the structure corresponded to
     {\em Craige's mom likes her}  then the NP for {\em Craige\/}  would be an
NP to the left of $p$ that
     has an NP node between it and $X$, and {\em Craige\/} would be selected as
the antecedent for {\em her}.

 \item
 The node $X$ is the highest S node in $\rm{U_2}$, so we go to the previous
sentence $\rm{U_1}$.
 As we traverse the tree of $\rm{U_1}$, the first NP we encounter is
$\rm{NP_1}$, so {\em Lyn's mom}
 is proposed as the antecedent for {\em her} and we are done.

\end{enumerate}

\end{document}